\documentclass[journal]{IEEEtran}
\usepackage{tikz,amsmath, amssymb,bm,color}
\usetikzlibrary{arrows}
\usepackage{graphicx}
\usepackage{float}
\usepackage{multirow}
\usepackage{tikz}
\usepackage{pgfplots}
\usepackage{pgfplotstable}
\usepackage{flushend}
\usepackage{hyperref}
\pgfplotsset{compat=1.14}
\ifCLASSINFOpdf
\else
\fi
\begin{document}
%
\title{Backward-Forward Algorithm: An Improvement towards Extreme Learning Machine}
%
%
%

\author{Dibyasundar~Das,
        Deepak~Ranjan~Nayak,
        Ratnakar~Dash,
        and~Banshidhar~Majhi
\thanks{Dibyasundar~Das, Deepak~Ranjan~Nayak,  Ratnakar~Dash and Banshidhar~Majhi is with the Department
of Computer Science and Engineering, National Institute of Technology Rourkela,
Odisha, India, 769008 e-mail: (dibyasundar@ieee.org).\newline github link: \href{https://github.com/Dibyasundar/BackwardForwardELM}{https://github.com/Dibyasundar/BackwardForwardELM}}
\thanks{~}}

%
%

\markboth{~}
{Shell \MakeLowercase{\textit{et al.}}: Bare Demo of IEEEtran.cls for Journals}
%



\maketitle

\begin{abstract}
The extreme learning machine needs a large number of hidden nodes to generalize a single hidden layer neural network for a given training data-set. The need for more number of hidden nodes suggests that the neural-network is memorizing rather than generalizing the model. Hence, a supervised learning method is described here that uses Moore-Penrose approximation to determine both input-weight and output-weight in two epochs, namely, backward-pass and forward-pass. The proposed technique has an advantage over the back-propagation method in terms of iterations required and is superior to the extreme learning machine in terms of the number of hidden units necessary for generalization.
\end{abstract}

\begin{IEEEkeywords}
Extreme Learning Machine, Single Layer Feed-forward Network , Image classification.
\end{IEEEkeywords}

%
\IEEEpeerreviewmaketitle

\section{Introduction}
\IEEEPARstart{M}{achine} learning is one of the key element to many of the real-world applications like text recognition \cite{ref1,ref4,ref72}, speech recognition \cite{fu2019applications,Lokesh2019}, automated CAD system \cite{NAYAK2018232, BEURA20151, MISHRA2019303}, defense\cite{KILMER199691, 8709752}, industry \cite{SUN2019257, HAN2019}, behavioral analysis\cite{pereira2019fast}, marketing\cite{gupta2019comparative} etc. Among the learning models, the neural network is well known for its flexibility in the choice of architecture and approximation ability. The Single hidden feed-forward neural network (SLFN) architecture is a widely used model for handling prediction and pattern classification problems. However, the weight and bias of these neural networks are popularly tuned using the gradient-based optimizer. These methods are known to be slow due to the improper choice of learning rate and may converge to local minima. Moreover, the learning iterations add computational cost to the tuning process of the model. Oppose to traditional methods randomized algorithms for training single layer feed-forward neural networks such as extreme learning machine (ELM)~\cite{ref90} and radial basis function network (RBFN)~\cite{broomhead1988radial}, have become a popular choice in recent years because of their generalization capability with faster learning speed ~\cite{cui2018elm,liu2018extreme,song2018segmentation}. Huang et al. \cite{ref90} have proposed ELM that takes advantage of the random transformation of the input feature to learn a generalized model in one iteration. In this method, the input weight and bias are chosen randomly for a given SLFN architecture, and output weight is analytically determined with the generalized inverse operation.

On the other hand, RBFN uses distance-based random feature mapping (centers of RBFs are generated randomly). However, RBFN obtains an unsatisfactory solution for some cases and results in poor generalization~\cite{wang2016randomized}. Hence, ELM provides effective solution for SLFNs with good generalization and extreme fast leaning, thereby, has been widely applied in various applications like regression~\cite{ref94}, data classification~\cite{ref90,ref94}, image segmentation~\cite{pan2012leukocyte}, dimension reduction~\cite{kasun2016dimension}, medical image classification~\cite{xie2016breast,zhang2017smart,nayak2017discrete},  face classification~\cite{mohammed2011human},  etc.  In~\cite{ref94}, Huang et al. discussed the universal approximation capability and scalability of ELM. The accuracy of classification in ELM depends on the choice of weight initialization scheme and activation function. To overcome this shortcoming, many researchers have used optimization algorithms that choose the best weight for the input layer. However, with the introduction of heuristic optimization, the choice of iteration and hyper-parameters are again introduced. Hence, such methods suffer from the same problem as the back-propagation based neural network. Thus here, we propose a non-iterative and non-parametric method that overcome the limitations of ELM and iterative-ELM. The main contribution of the paper is to develop a non-iterative and non-parametric algorithm, namely backward-forward ELM, to train a single hidden layer neural network.

A comprehensive study of the proposed model on many of standard machine learning classification and prediction applications. As well as, two well know image classification data-sets, namely MNIST and Brain-MRI, are studied for non-handcrafted feature evaluation. The rest of the paper is organized as follows. Section II gives an overview of the motivation and objective behind the development of the ELM algorithm and its limitations. In the next section, the proposed backward-forward ELM algorithm is described in brief. Section IV summarizes the experiments conducted, and finally, Section V concludes the study.

\section{Extreme Learning Machine}
 Feed-forward Neural network is slow due to gradient-based weight learning and the requirement of parameter tuning. The extreme learning machine is one of the learning models for the single hidden feed-forward neural network (SLFN) where the input-weights are randomly chosen, and the output-weights are determined analytically. This makes the network to converge to the underlying regression layer in one pass, which is a faster learning algorithm than the traditional gradient-based algorithms. The development of the ELM algorithm is based on the assumption that input weight and bias do not create much difference in obtained accuracy, and a minimum error is acceptable if many computational steps can be avoided. However, the accuracy and generalization capability highly depends on the learning of the output-weight and minimization of output-weight norm.
 
 \begin{table}
 	\centering
 	\caption{List of symbols}
 	\renewcommand{\arraystretch}{1.3}
 	\resizebox{0.49\textwidth}{!}{
 	\begin{tabular}{ll}
 		\hline 
 		Symbol & Meaning\\
 		\hline 
 		$N$ & Number of samples\\
 		$P$ & Size of input nodes\\
 		$M$ & Size of hidden nodes\\
 		$C$ & Size of output nodes\\
 		$x_{j}$ & Input vector $[x_{(j,1)}, x_{(j,2)}, \ldots, x_{(j,P)}]^T$ where, $j = 1,2, \ldots, N$\\
 		$I$&Augmented input data-set\\
 		& $I=\left[\begin{array}{lllll}
 		x_{(1,1)}& x_{(1,2)}& \ldots & x_{(1,P)}& 1\\
 		x_{(2,1)}& x_{(2,2)}& \ldots & x_{(2,P)}& 1\\
 		\vdots& \vdots& \vdots & \vdots&\vdots\\
 		x_{(N,1)}& x_{(N,2)}& \ldots & x_{(N,P)}& 1\\
 		\end{array}\right]$\\
 		$t_{j}$ & Output vector $[t_{(j,1)}, t_{(j,2)}, \ldots, t_{(j,C)}]^T$ where, $j = 1,2, \ldots, N$\\
 		$w_{i}$ & Input weight $[w_{(i,1)}, w_{(i,2)}, \ldots, w_{(i,P)}]$ where, $i = 1,2, \ldots, M$\\
 		$b_{i}$ & Input bias where, $i = 1,2, \ldots, M$\\
 		$W$&Input weight\\
 		& $W=\left[\begin{array}{lllll}
 		w_{(1,1)}& w_{(1,2)}& \ldots & w_{(1,P)}& b_1\\
 		w_{(2,1)}& w_{(2,2)}& \ldots & w_{(2,P)}& b_2\\
 		\vdots& \vdots& \vdots & \vdots&\vdots\\
 		x_{(M,1)}& x_{(M,2)}& \ldots & w_{(M,P)}& b_M\\
 		\end{array}\right]^T$\\
 		$\beta_{k}$ & Output weight $[\beta_{(k,1)}, \beta_{(k,2)}, \ldots, \beta_{(k,M)}]$ where, $k = 1,2, \ldots, C$\\
 		$g(.)$ & Activation function\\
 		$( )^{\dagger}$ & Pseudoinverse\\
 		$ortho(.)$ & Orthogonal transformation\\
 		\hline 
 	\end{tabular}
	}
 \end{table}

  The approximation problem can be expressed as follows;

For \(N\) distinct samples \((x_j,t_j)\), \(M\) hidden neurons and \(g(.)\) be the activation function, so the output of SLFN can be modeled as:
\begin{equation}
o_j = \sum_{i=1}^{M}\beta_{i} . g(w_i . x_j + b_i)  \textrm{  ,  for } j = 1,\ldots,N
\end{equation}
Hence, the error ($E$) for the target output ($t$) is \(\sum_{j=1}^{N}||o_j-t_j||\) and it can be expressed as;
\begin{equation}
E=||\sum_{j=1}^{N}(\sum_{i=1}^{M}\beta_{i}. g(w_i . x_j + b_i) )- t_j||
\end{equation}

For an ideal approximation case error is zero. Hence,

\begin{equation}
\begin{array}{llll} &||\sum_{j=1}^{N}(\sum_{i=1}^{M}\beta_{i}. g(w_i . x_j + b_i) )- t_j||&=&0 \\
&&&\\ 
\Rightarrow & \sum_{i=1}^{M}\beta_{i} g(w_i . x_j + b_i) &=& t_j\\ &\textrm{ for all } j=1, \dots,N\\ \end{array}
\end{equation}

This equation can be expressed as

\begin{equation}
H\beta=T
\end{equation}

where,

\begin{equation}
\begin{array}{l}
\begin{array}{c|ccc|}
&g(w_1 . x_1 + b_1) & \ldots &g(w_M . x_1 + b_M)\\
&.&\ldots&.\\
H=&.&\ldots&.\\
&.&\ldots&.\\
&g(w_1 . x_N + b_1) & \ldots &g(w_M . x_N + b_M)\\
\end{array}\\\\
\begin{array}{c|c|}
&\beta_1\\
&.\\
\beta=&.\\
&.\\
&\beta_M\\
\end{array} \textrm{ and}
\begin{array}{c|c|}
&t_1\\
&.\\
T=&.\\
&.\\
&t_N\\
\end{array}
\end{array}
\end{equation}

If given N==M (i.e., the sample size is the same as the number of the hidden neurons); the matrix H is square and invertible if its determinant is nonzero. In such a case, the SLFN can approximate with zero error. But in reality, \(M<<N\) hence \(\beta\) is not invertible. Hence rather finding an exact solution, we try to find a near-optimal solution that minimizes the approximation error. Which can be expressed as;
\begin{equation}
||\hat{H}\hat{\beta}-T|| \simeq ||H\beta-T||
\end{equation}

\(\hat{H}\) and \(\hat{\beta}\) can be defined as
\begin{equation}
\begin{array}{l}
\begin{array}{c|ccc|} &\hat{g}(\hat{w}_1 . x_1 + \hat{b}_1) & \ldots &\hat{g}(\hat{w}_M . x_1 + \hat{b}_M)\\ &.&\ldots&.\\ \hat{H}=&.&\ldots&.\\ &.&\ldots&.\\ &\hat{g}(\hat{w}_1 . x_N + \hat{b}_1) & \ldots &\hat{g}(\hat{w}_M . x_N + \hat{b}_M)\\ \end{array}\\\\ \textrm{, and} \begin{array}{c|c|} &\hat{\beta}_1\\ &.\\ \hat{\beta}=&.\\ &.\\ &\hat{\beta}_M\\ \end{array}
\end{array}
\label{eq_H}
\end{equation}

In any learning method for SLFN we try to find $\hat{w}$, $\hat{b}$, $\hat{g}(.)$ and $\hat{\beta}$ in order to minimize the error of prediction. Mostly $\hat{g}(.)$ is chosen as a continuous function depending on the model consideration of data (various activation functions are Sigmoid, tan-hyperbolic, ReLU, etc.). The $\hat{w}$, $\hat{b}$ and $\hat{\beta}$ are to be determined by the learning algorithm. Back-propagation is one of the most famous learning algorithms that use the gradient descent method. However, the gradient-based algorithms have the following issues associated with them:

\begin{enumerate}
	\item
	Choosing proper learning rate \(\eta\) value. Small \(\eta\) converges very slowly, and Very high value of \(\eta\) makes the algorithm unstable.
	\item
	The gradient-based learning some times may converge to local minima, which is undesirable if the difference between global minima and local minima is significantly large.
	\item
	Some times overtraining leads to worse generalization, hence proper stopping criteria are also needed.
	\item
	Gradient-based learning is very time-consuming.

\end{enumerate}

For above reasons the ELM chooses $\hat{w}$, $\hat{b}$ randomly and uses MP inverse to calculate \(\hat{\beta}\) analytically. Hence \(\hat{\beta}\) can be expressed as

\begin{equation}
\hat{\beta}=\hat{H}^{\dagger}.T=(\hat{H}' . \hat{H})^{-1}.\hat{H}' .T
\label{eq_beta}
\end{equation}

\noindent \textbf{Drawbacks of ELM:}\\

Das et al. \cite{Das2019} have studied deeply on the behavior of the ELM, for various weight initialization schemes, activation functions, and the number of nodes. From this study, it is found that ELM has limitations as follows.

\begin{itemize}
	\item The accuracy of classification in ELM depends on the choice of weight initialization scheme and activation function.
	\item 
	It is observed that the ELM needs relatively higher hidden nodes to provide higher accuracy. The need for more hidden nodes, suggests the network is memorizing the samples rather than providing a generalized performance.
	\item 
	It is also observed that due to random weights in the final network, ELM suffers from ill-posed problems.
\end{itemize}

 To overcome these shortcomings, many researchers have used optimization algorithms \cite{NAYAK2019105626, NAYAK2018232}, which choose the best weight for the input layer. However, such a solution again introduces the iteration and choice of parameter problem for the optimization scheme. Hence,  this paper proposes a backward-forward method for a single hidden layer neural network which has the following advantages over other learning models:
 \begin{itemize}
 	\item The algorithm generalizes the network with few hidden layer nodes only in two steps. In, the first step (backward pass), the input weights are evaluated, and in the second step (forward pass), the suitable output weight is determined.
 	\item The final model of the network does not contain any random weights, thus giving a consistent result even when the choice of activation changes.
 	\item 
 	Unlike optimization-based ELM, the proposed method evaluates input weight in two steps. Hence the model does not need iterative steps.
 \end{itemize} 
\section{Proposed backward forward algorithm for ELM}
In this section, we discussed the learning process of the proposed model. In the architecture of a single hidden layer neural network, there are two types of weights to learn, namely input-weight (weight matrix that represents connection from input to hidden layer) and output-weight (weight matrix that represent connection from hidden to output layer). The proposed model has two major stages, namely backward-pass (where input weights are learned), and forward-pass (where output weights are determined). We made the following assumption to develop the proposed backward forward algorithm for ELM (BF-ELM) algorithm.

\begin{itemize}
	\item The weights in the neural network can be categories into two parts. Some of the weights generalize the model, and the rest of the weights is used to memorize the samples. Hence, in backward-pass, the BF-ELM determines half of the weights that are assumed to generalize the model for a given training data-set. 
	\item If a learned model uses linear activation and the activation is replaced, it will not affect the accuracy of the model. Hence, in backward-pass, the model assumes linear activation, and proper activation is replaced in forward-pass.
	\item If the input training set ($I$) and hidden layer output ($H$) is augmented, then bias can be ignored.
\end{itemize}

Both of the stages are described in detail in the following sections. 
\subsection{Backward-pass}
In backward-pass the model learns a subset of input-weight using Moore-Penrose inverse in direction from output to input. For, a given a training set $\mathcal{N}=\{(x_j,t_j)~|~x_j\epsilon\mathbb{R}^P,~t_j\epsilon\mathbb{R}^C,~\text{where }j=1,2,\ldots,N\}$ we design a SLFN with $M/2$ hidden nodes which determines a subset of input-weight ($\widetilde{W}\text{ of size }(P, M/2)$) as follows.
\begin{enumerate}
	\item The output-weight {$\widetilde{\beta}$} of size($M/2,c$) is set randomly.
	\item  The hidden layer output matrix is determined using following equation.
	\begin{equation}
	\widetilde{H}= T \times \widetilde{\beta}^{\dagger} + \mathrm{Random~Error}
	\label{n_eq1}
	\end{equation}
	\item The subset of input-weight($W$) is determined by following equation.
	\begin{equation}
	\widetilde{W}= I^{\dagger}*\widetilde{H}
	\label{n_eq2}
	\end{equation}
	\item The learned subset ($\widetilde{W}$) is used to determine full input-weight ($W\text{ of size }(P, M)$)by appending orthogonal transformation of $\widetilde{W}$ as follows.
	\begin{equation}
	W = \left[\widetilde{W}, \mathrm{orth}(\widetilde{W})\right]
	\label{n_eq3}
	\end{equation}
\end{enumerate}
\subsection{Forward-pass}
in next stage the learned input-weight ($W$) is used to find output-weight ($\hat{\beta}$) is determined in forward-pass.
\begin{enumerate}
	\item The hidden-layer is determined by using $W$.
	 \begin{equation}
	 H=g(I\times W)
	 \label{n_eq4}
	 \end{equation}
	 where, g(.) is the activation function.
	 \item Finally, the output-weight is determined as follows:
	 \begin{equation}
	 \beta=H^{\dagger}\times T
	 \label{n_eq5}
	 \end{equation}
\end{enumerate}

\begin{figure}[H]
	\centering
	\resizebox{0.4\textwidth}{!}{
		\begin{tikzpicture}
		
		\draw[fill=red!20]  (-12.5,7) rectangle (3,-1);
		\draw[fill=green!20]  (-12.5,-1) rectangle (3,-9.5);
		\pgfsetlinewidth{0.8}
		\draw  (-11,4) ellipse (0.3 and 0.3);
		\draw  (-11,3) ellipse (0.3 and 0.3);
		\draw  (-11,1) ellipse (0.3 and 0.3);
		\node[rotate=90] at (-11,2) {\Huge $\ldots$};
		\draw  (-11.5,4.5) rectangle (-10.5,0.5);
		
		\draw  (-8,5) ellipse (0.3 and 0.3);
		\draw  (-8,4) ellipse (0.3 and 0.3);
		\draw  (-8,0) ellipse (0.3 and 0.3);
		\draw  (-8,3) ellipse (0.3 and 0.3);
		\draw  (-8,2) ellipse (0.3 and 0.3);
		\node[rotate=90] at (-8,1) {\Huge $\ldots$};
		\draw  (-8.5,5.5) rectangle (-7.5,-0.5);

		\draw[-triangle 45] (-8.5,2.5) -- (-9.5,2.5);
		\draw[-triangle 45] (-10.5,2.5) -- (-9.5,2.5);

		\node[scale=1.5] at (-9.5,3.5) {$\widetilde{W}=?$};
		
		\draw  (-5,6) ellipse (0.3 and 0.3);
		\draw  (-5,3) ellipse (0.3 and 0.3);
		\draw  (-5,5) ellipse (0.3 and 0.3);
		\node[rotate=90] at (-5,4) {\Huge $\ldots$};
		\draw  (-5.5,6.5) rectangle (-4.5,2.5);
		
		\draw  (-6,1) rectangle (-4,0);
		\draw  (-6.5,2.5) ellipse (0.5 and 0.5);
		\draw[-triangle 45]  (-7,2.5) -- (-7.5,2.5);
		\draw[-triangle 45]  (-5.5,4.5) -- (-6.5,4.5) -- (-6.5,3);
		\draw[-triangle 45]  (-6,0.5) -- (-6.5,0.5) -- (-6.5,2);
		\node at (-6.5,2.5) {\Huge $+$};
		\node[scale=1.5] at (-5,0.5) {$Error$};
		\node[scale=1.5]  at (-6,5) { $\widetilde{\beta}$};
		\node[anchor=south, ,scale=1.5] at (-8,5.5) {$\widetilde{H}$};
		\draw[-triangle 45] (-4,4.5) -- (-4.5,4.5);
		\node[anchor=west, ,scale=1.5] at (-4,4.5) {$T$};
		\node[anchor=west, ,scale=1.5] at (-3,3.5) {$\widetilde{H} = T*\widetilde{\beta}^{\dag} + Error$};
		\draw[-triangle 45] (-12,2.5) -- (-11.5,2.5);
		\node[anchor=east, scale=1.5] at (-12,2.5) {$I$};
		
		\draw[-triangle 45,line width=2.5] (3,7.5) -- (-12.5,7.5);
		\node[scale=1.5] at (-5.5,8) {\bf Backward Pass};
		\node[anchor=west, scale=1.5] at (-3,2.5) {$W_1=I^{\dag}*H'$};
		
		\node[anchor=west, scale=1.4] at (-3,1.5) {\bf $\dag : $ Pseudoinverse };

		\draw  (-11,-4.5) ellipse (0.3 and 0.3);
		\draw  (-11,-5.5) ellipse (0.3 and 0.3);
		\draw  (-11,-7.5) ellipse (0.3 and 0.3);
		\node[rotate=90] at (-11,-6.5) {\Huge $\ldots$};
		\draw  (-11.5,-4) rectangle (-10.5,-8);

		\draw  (-8,-3.5) ellipse (0.3 and 0.3);
		\draw  (-8,-4.5) ellipse (0.3 and 0.3);
		\draw  (-8,-8.5) ellipse (0.3 and 0.3);
		\draw  (-8,-5.5) ellipse (0.3 and 0.3);
		\draw  (-8,-6.5) ellipse (0.3 and 0.3);
		\node[rotate=90] at (-8,-7.5) {\Huge $\ldots$};
		\draw  (-8.5,-3) rectangle (-7.5,-9);
		
		\draw  (-5,-4.5) ellipse (0.3 and 0.3);
		\draw  (-5,-7.5) ellipse (0.3 and 0.3);
		\draw  (-5,-5.5) ellipse (0.3 and 0.3);
		\node[rotate=90] at (-5,-6.5) {\Huge $\ldots$};
		\draw  (-5.5,-4) rectangle (-4.5,-8);
		
		\draw[-triangle 45, line width=2.5] (-12.5,-10) -- (3,-10);
		\node[scale=1.5] at (-5.5,-10.5) {\bf Forward pass};
		\draw[-triangle 45] (-10.5,-6) -- (-8.5,-6);
		\draw[-triangle 45] (-12,-6) -- (-11.5,-6);
		\draw[-triangle 45] (-7.5,-6) -- (-5.5,-6);
		\draw[-triangle 45] (-4.5,-6) -- (-4,-6);
		\draw[-triangle 45] (-10,3) .. controls (-10,0.5) and (-10,0.5) .. (-10,-1.75);
		
		\node[scale=1.5] at (-9.5,-5.5) {$W$};
		\node[anchor=east, scale=1.5] at (-12,-6) {$I$};
		
		\node[scale=1.5] at (-6.5,-5.5) {$\beta=?$};
		\node[anchor=west, scale=1.5] at (-4,-6) {$T$};

		\node[anchor=south,scale=1.5] at (-8,-3) {$H$};
		\node[anchor=west, scale=1.5] at (-3,-3.5) {$H= \mathrm{g}(I \times W )$};
		\node[anchor=west, scale=1.5] at (-3,-4.5) {$\beta = H^{\dag}*T$};
		
		\draw[line width=1.5] (-12.5,-1) -- (3,-1);
		\node[anchor=west,scale=1.5] at (-3,-5.5) {\bf $\mathrm{g}(.)$ : Activation };
		\node[scale=1.5] at (-9.5,-2) {$W=[\widetilde{W}, \mathrm{ortho}(\widetilde{W})]$};
		
		\draw[-triangle 45]  (-9.5,-2.5) .. controls (-10.5,-3.5) and (-9,-3.5) .. (-9.5,-5);
		\node[anchor=west,scale=1.5] at (-3,-6.5) {\bf $\mathrm{ortho}(.)$ : Orthogonal};
		\node[anchor=west,scale=1.5] at (0,-7) {\bf  Transform};
		\end{tikzpicture}
	
	}
	\caption{The proposed backward forward extreme Learning Machine (BF-ELM)}\label{model}
\end{figure}
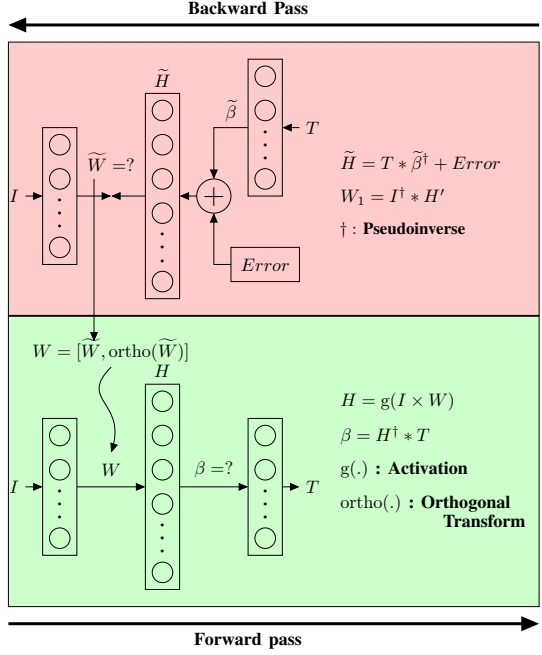

The over all diagram of the proposed BF-ELM model is given in Fig. \ref{model}, which shows the determination of input weight ($W$) and output weight($\beta$). In, next section various experiments have been carried out on multiple image classification data-set that shows the learning capability of the BF-ELM. The proposed algorithm needs fewer number of nodes compared to ELM to achieve better generalization performance.

\section{Performance evaluation}
In this section performance of the proposed BF-ELM is compared with ELM on various benchmark data-sets. The comparison is made with respect to the number of hidden neurons required for generalized performance and the time needed to compute the output for the testing set. All implementations of BF-ELM and ELM are carried out in MATLAB 2018b running on an i7-4710HQ processor with Ubuntu operating system. The pseudoinverse ($^\dagger$) operation is done using MATLAB in-built function, and the ELM implementation is done following the paper~\cite{ref90}. The experiment conducted can be divided in two-part; the first experiment compares the two algorithms on the basis of hidden nodes required, and the second experiment observes the behavior of models with respect to change in weight initialization scheme and activation function as described in TABLE  \ref{tab_weight} and \ref{tab_activation} respectively. The test is conducted for each combination of weight initialization scheme and activation function. 
\begin{table}[H]
	\caption{Weight initialization scheme investigated in this work}\label{tab_weight}
	\centering
	\renewcommand{\arraystretch}{1.5}
	\begin{tabular}{p{4cm}p{4cm}}
		\hline
		Name&Description\\\hline
		Uniform random initialization & $W\sim U[l,u]$, where, $l$ represents lower range and $u$ represents upper range of the uniform distribution $U$\\\hline
		Xavier initialization  & $
		W\sim N\left(0,\frac{2}{n_{in}+n_{out}}\right)$	where $n_{in}$ $n_{out}$ represent the input layer size (dimension of features) and the output layer size (number of classes) respectively.\\\hline
		ReLU initialization& $	W\sim N\left(0,\sqrt{\frac{2}{n_c}}\right)$ where, $n_{c}$ is hidden nodes size\\\hline
		Orthogonal initialization& Random orthogonal matrix each row with orthogonal vector\\\hline
	\end{tabular}
\end{table}
\begin{table}[H]
	\caption{Activation functions investigated in this work}\label{tab_activation}
	\centering
	\renewcommand{\arraystretch}{1.5}
	\begin{tabular}{ll}
		\hline
		Activation function& Expression\\\hline
		Linear&$g(x)=x$\\
		
		Sigmoid&$g(x)=\frac{1}{1+e^{-x}}$\\
		
		ReLu&$
		g(x)=\left\{\begin{array}{ll}
		x&\mathrm{if~}x>0 \\
		0&\mathrm{if~}x\le 0 \\
		\end{array}\right.
		$\\
		Tanh
		&$
		g(x)=\frac{e^{x}-e^{-x}}{e^{x}+e^{-x}}
		$\\
		Softsign
		&$
		g(x)=\frac{1}{1+|x|}
		$\\
		Sin
		&$
		g(x)=sin(x)
		$\\
		
		Cos
		
		&$
		g(x)=cos(x)
		$\\
		
		Sinc
		
		&$
		g(x)=\left\{
		\begin{array}{ll}
		1& \mathrm{if ~}x=0\\
		\frac{sin(x)}{x}& \mathrm{if ~}x\neq 0\\
		\end{array}
		\right.
		$\\
		
		LeakyReLu
		
		&$
		g(x)=\left\{\begin{array}{ll}
		x&\mathrm{if~}x>0 \\
		0.001.x&\mathrm{if~}x\le 0 \\
		\end{array}\right.
		$\\
		
		Gaussian
		
		&$
		g(x)=e^{-x^2}
		$\\
		
		Bent Identity
		
		&$
		g(x)=\frac{\sqrt{x^2+1}-1}{2}+x
		$\\\hline
	\end{tabular}
\end{table}

The brief description of the benchmark data-sets and the result analysis are given as follows:
\subsection{Benchmark with sine cardinal regression problems}
~\\The approximation function sine cardinal (as given in equation \ref{ss}) is used to test the proposed learning model.First 5000 data points for training set are generated, where $x$ is randomly distributed over [-10,10] with additive random error of uniform distribution of [-0.2,0.2] to response $y$. The testing set is created without using any additive error.
\begin{equation}
	y(x)=\left\{
		\begin{array}{ll}
		sin(x)/x&x\neq 0\\
		1&x=0\\
		\end{array}
		\right.\label{ss}
\end{equation}

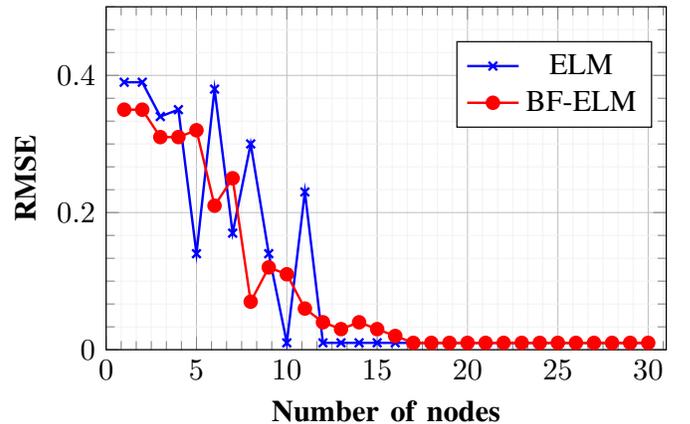
\begin{figure}[H]
	\resizebox{0.49\textwidth}{!}{\begin{tikzpicture}
		\begin{axis}[
		grid=both,
		grid style={line width=.1pt, draw=gray!10},
		major grid style={line width=.2pt,draw=gray!50},
		minor tick num=5,
		scale only axis, 
		height=4cm,
		width=0.36\textwidth,
		xmax=31, xmin=0,
		ymax=0.5, ymin=0, 
		legend pos=north east,
		xtick={0,5,10,15,20,25,30}, 
		xlabel={\textbf{Number of nodes}},
		ylabel={\textbf{RMSE}}, ylabel near ticks,
		legend style={
			at={(0.8,0.9)},
			anchor=north}
		]
		\addplot[blue, thick,mark=x] table [y=d,x=c,col sep=comma] {Resultsinc.txt};
		\addplot[red, thick,mark=*] table [y=f,x=c,col sep=comma] {Resultsinc.txt};
		\legend{ELM,BF-ELM}
		\end{axis}
		\end{tikzpicture}}
	\caption{Accuracy comparison of BF-ELM to ELM w.r.t number of nodes on sine cardinal regression problems}\label{Sinc_node}
\end{figure}

The experiment to analyze the hidden nodes required to solve the regression problem for both ELM and BF-ELM is done. During the experiment the activation is set to sin function. The obtained root-mean-squared-error (RMSE) is depicted in Fig. \ref{Sinc_node}.

\begin{figure}[H]
	\includegraphics[width=0.49\textwidth]{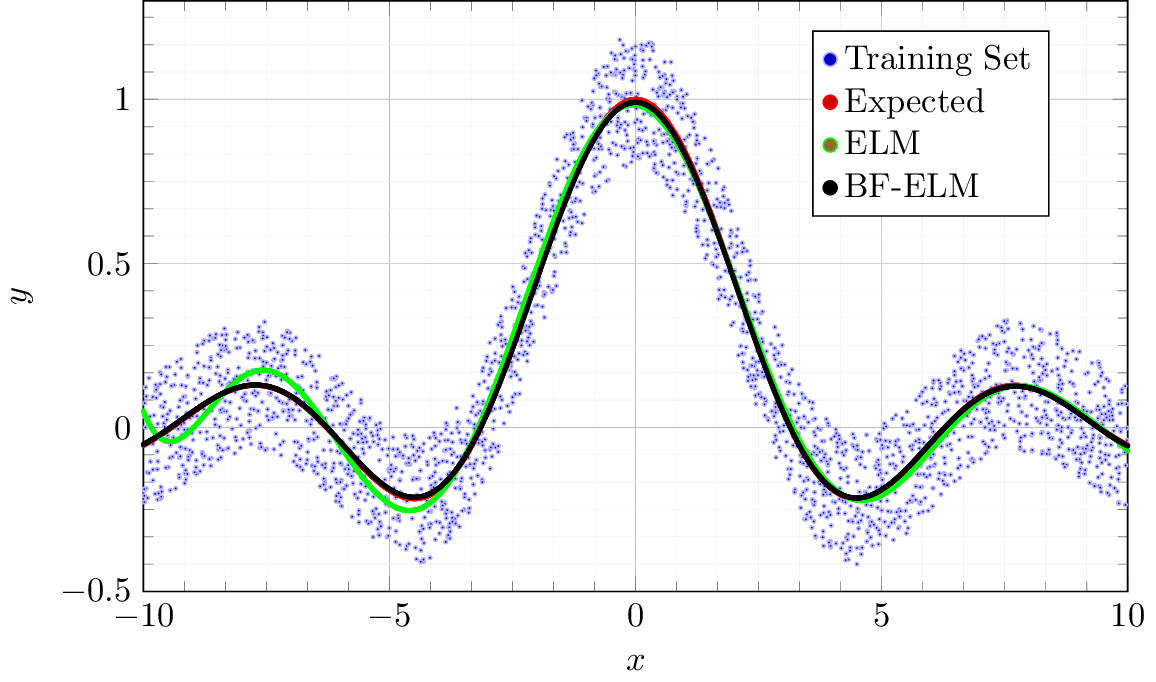}
	\caption{The comparison of ELM and BF-ELM for SLFN with 4 hidden nodes and sinc activation for approximation of \ref{ss}} \label{Sinc_ap}
\end{figure}

 The RMSE decreases while increasing the number of hidden nodes. It is observed that the BF-ELM minimizes error with less number of hidden nodes up to 12 nodes then ELM result are superior and equilibrium point is achieved with 17 or more hidden nodes. The effect of various activation function and weight initialization scheme is summarized in TABLE \ref{Sinc_app} with respect to root mean square error (RMSE) and testing time. The hidden nodes for both ELM and BF-ELM algorithms are set to 10. 
\begin{table}[H]
	\renewcommand{\arraystretch}{1.3}
	\centering
	\caption{RMSE comparison on sinc regression for 10 hidden nodes}\label{Sinc_app}
	\resizebox{0.49\textwidth}{!}{\begin{tabular}{llllll}
		\hline
		\textbf{Activation} 	&	\textbf{Weight} 	&	\multicolumn{2}{c}{\textbf{ELM}} 	&	\multicolumn{2}{c}{\textbf{BF ELM}} 	\\\cline{3-6}
		\textbf{function}	&	\textbf{initialization}	&	\textbf{RMSE}&\textbf{Test Time}	&	\textbf{RMSE}&\textbf{Test Time}	\\\hline
		\multirow{5}{*}{Relu}	&ortho &0.18	&0.000265	&0.14	&0.000245\\
		&rand(0,1) &0.21	&0.000264	&0.20	&0.000252\\
		&rand(-1,1) &0.10	&0.000260	&0.16	&0.000215\\
		&xavier &0.21	&0.000277	&0.21	&0.000232\\
		&relu &0.09	&0.000270	&0.22	&0.000304\\\hline
		\multirow{5}{*}{Sigmoid}	&ortho &0.02	&0.000305	&0.03	&0.000297\\
		&rand(0,1) &0.03	&0.000298	&0.04	&0.000296\\
		&rand(-1,1) &0.02	&0.000304	&0.03	&0.000293\\
		&xavier &0.05	&0.000315	&0.06	&0.000294\\
		&relu &0.01	&0.000296	&0.02	&0.000297\\\hline
		\multirow{5}{*}{Tanh}	&ortho &0.03	&0.000404	&0.04	&0.000392\\
		&rand(0,1) &0.06	&0.000407	&0.07	&0.000392\\
		&rand(-1,1) &0.06	&0.000410	&0.03	&0.000401\\
		&xavier &0.06	&0.000410	&0.06	&0.000402\\
		&relu &0.01	&0.000395	&0.06	&0.000395\\\hline
		\multirow{5}{*}{Softsign}	&ortho &0.09	&0.000153	&0.13	&0.000148\\
		&rand(0,1) &0.12	&0.000148	&0.13	&0.000138\\
		&rand(-1,1) &0.11	&0.000145	&0.11	&0.000140\\
		&xavier &0.07	&0.000144	&0.06	&0.000131\\
		&relu &0.08	&0.000144	&0.13	&0.000137\\\hline
		\multirow{5}{*}{Sin}	&ortho &0.03	&0.000292	&0.12	&0.000234\\
		&rand(0,1) &0.01	&0.000289	&0.10	&0.000238\\
		&rand(-1,1) &0.01	&0.000307	&0.03	&0.000232\\
		&xavier &0.12	&0.000291	&0.04	&0.000192\\
		&relu &0.04	&0.000176	&0.04	&0.000184\\\hline
		\multirow{5}{*}{Cos}	&ortho &0.03	&0.000213	&0.03	&0.000201\\
		&rand(0,1) &0.01	&0.000236	&0.03	&0.000199\\
		&rand(-1,1) &0.01	&0.000240	&0.10	&0.000182\\
		&xavier &0.24	&0.000257	&0.09	&0.000179\\
		&relu &0.02	&0.000243	&0.02	&0.000197\\\hline
		\multirow{5}{*}{Sinc}	&ortho &0.08	&0.000492	&0.01	&0.000444\\
		&rand(0,1) &0.01	&0.000565	&0.03	&0.000655\\
		&rand(-1,1) &0.01	&0.000702	&0.03	&0.000594\\
		&xavier &0.02	&0.000661	&0.08	&0.000606\\
		&relu &0.01	&0.000687	&0.08	&0.000496\\\hline
		\multirow{5}{*}{BentIde}	&ortho &0.03	&0.000169	&0.03	&0.000174\\
		&rand(0,1) &0.08	&0.000170	&0.04	&0.000166\\
		&rand(-1,1) &0.03	&0.000183	&0.07	&0.000166\\
		&xavier &0.08	&0.000171	&0.03	&0.000168\\
		&relu &0.02	&0.000178	&0.03	&0.000168\\\hline
		\multirow{5}{*}{ArcTan}	&ortho &0.06	&0.000212	&0.11	&0.000191\\
		&rand(0,1) &0.06	&0.000237	&0.05	&0.000208\\
		&rand(-1,1) &0.08	&0.000235	&0.08	&0.000189\\
		&xavier &0.02	&0.000220	&0.06	&0.000205\\
		&relu &0.02	&0.000198	&0.03	&0.000204\\\hline
	\end{tabular}}
\end{table}
TABLE \ref{Sinc_app} shows the minimum RMSE value in every weight initialization and activation function combination. As the architecture for SLFN remains constant for both ELM and BF-ELM, hence, the testing time for both algorithms are nearly similar. Fig. \ref{Sinc_ap} shows the approximated function learned by ELM and BF-ELM for the input training data. The best result was obtained with four hidden nodes and sinc activation. BF-ELM learned the approximated values nearly to actual expected value of generalization. 

\subsection{Benchmark with iris data-set}

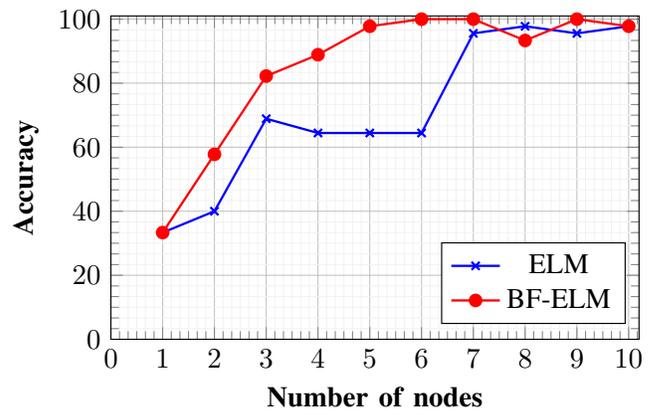
\begin{figure}[H]
	\resizebox{0.49\textwidth}{!}{
		\begin{tikzpicture}
		\begin{axis}[
		grid=both,
		grid style={line width=.1pt, draw=gray!10},
		major grid style={line width=.2pt,draw=gray!50},
		minor tick num=5,
		scale only axis, 
		height=4cm,
		width=0.36\textwidth,
		xmax=10.2, xmin=0,
		ymax=101, ymin=0, 
		legend pos=north east,
		xtick={0,1,2,3,4,5,6,7,8,9,10}, 
		xlabel={\textbf{Number of nodes}},
		ylabel={\textbf{Accuracy}}, ylabel near ticks,
		legend style={
			at={(0.8,0.3)},
			anchor=north}
		]
		\addplot[blue, thick,mark=x] table [y=d,x=c,col sep=comma] {ResultIris.txt};
		\addplot[red, thick,mark=*] table [y=f,x=c,col sep=comma] {ResultIris.txt};
		\legend{ELM,BF-ELM}
		\end{axis}
		\end{tikzpicture}
	}
	\caption{Accuracy comparison of BF-ELM to ELM w.r.t number of nodes on iris dataset }\label{iris_node}
\end{figure}

The iris data-set use multiple measurements like sepal length, sepal width, petal length, petal width to classify taxonomy of 3 different species of iris namely Setosa, Versicolor, and Virginica. The data-set contains 50 samples per each class and the data-set is divided in to 70:30 training and testing set respectively. The accuracy increases with number of hidden-node and the performance comparison of BF-ELM and ELM with respect to number of hidden nodes is given in Fig. \ref{iris_node}.

The best accuracy for BF-ELM is obtained with six hidden nodes with orthogonal weight initialization and sigmoid activation function. Hence, further analysis of choice of weight initialization method and activation function is done with six hidden nodes. The summary of the analysis is given in TABLE \ref{iris_tab}.

\begin{table}[H]
	\renewcommand{\arraystretch}{1.3}
	\centering
	\caption{Accuracy comparison on iris data-set for 6 hidden nodes}\label{iris_tab}
	\resizebox{0.49\textwidth}{!}{\begin{tabular}{llrlrl}
			\hline
			\textbf{Activation} 	&	\textbf{Weight} 	&	\multicolumn{2}{c}{\textbf{ELM}} 	&	\multicolumn{2}{c}{\textbf{BF ELM}} 	\\\cline{3-6}
			\textbf{function}	&	\textbf{initialization}	&	\textbf{Acc}&\textbf{Test Time}	&	\textbf{Acc}&\textbf{Test Time}	\\\hline
			\multirow{5}{*}{None}	&ortho 	&77.78	&0.000038	&77.78	&0.000121\\
			&rand(0,1) 	&77.78	&0.000037	&77.78	&0.000087\\
			&rand(-1,1) 	&77.78	&0.000040	&77.78	&0.000121\\
			&xavier 	&77.78	&0.000036	&80.00	&0.000123\\
			&relu 	&77.78	&0.000029	&77.78	&0.000068\\\hline
			\multirow{5}{*}{Relu}	&ortho  &6.67	&0.000016	&80.00	&0.000062\\
			&rand(0,1) 	&77.78	&0.000020	&75.56	&0.000062\\
			&rand(-1,1) 	&82.22	&0.000013	&80.00	&0.000060\\
			&xavier  &6.67	&0.000022	&86.67	&0.000089\\
			&relu 	&97.78	&0.000014	&95.56	&0.000063\\\hline
			\multirow{5}{*}{Sigmoid}	&ortho 	&88.89	&0.000030	&100.00	&0.000143\\
			&rand(0,1) 	&77.78	&0.000027	&97.78	&0.000097\\
			&rand(-1,1) 	&100.00	&0.000032	&100.00	&0.000123\\
			&xavier 	&91.11	&0.000025	&100.00	&0.000157\\
			&relu 	&88.89	&0.000030	&100.00	&0.000105\\\hline
			\multirow{5}{*}{Tanh}	&ortho 	&97.78	&0.000031	&97.78	&0.000121\\
			&rand(0,1) 	&71.11	&0.000038	&100.00	&0.000110\\
			&rand(-1,1) 	&82.22	&0.000036	&100.00	&0.000111\\
			&xavier 	&100.00	&0.000040	&95.56	&0.000114\\
			&relu 	&80.00	&0.000039	&97.78	&0.000110\\\hline
			\multirow{5}{*}{Softsign}&ortho 	&77.78	&0.000022	&93.33	&0.000090\\
			&rand(0,1) 	&82.22	&0.000022	&93.33	&0.000090\\
			&rand(-1,1) 	&100.00	&0.000020	&100.00	&0.000090\\
			&xavier 	&97.78	&0.000020	&97.78	&0.000090\\
			&relu 	&88.89	&0.000021	&95.56	&0.000087\\\hline
			\multirow{5}{*}{Sin}	&ortho 	&93.33	&0.000036	&100.00	&0.000108\\
			&rand(0,1) 	&88.89	&0.000037	&100.00	&0.000114\\
			&rand(-1,1) 	&97.78	&0.000032	&97.78	&0.000120\\
			&xavier 	&100.00	&0.000355	&93.33	&0.000153\\
			&relu 	&100.00	&0.000202	&100.00	&0.000111\\\hline
			\multirow{5}{*}{Cos}	&ortho 	&97.78	&0.000058	&95.56	&0.000120\\
			&rand(0,1) 	&88.89	&0.000040	&95.56	&0.000109\\
			&rand(-1,1) 	&91.11	&0.000036	&100.00	&0.000111\\
			&xavier 	&95.56	&0.000038	&100.00	&0.000105\\
			&relu 	&86.67	&0.000037	&84.44	&0.000113\\\hline
			\multirow{5}{*}{LeakyRelu}	&ortho 	&93.33	&0.000036	&80.00	&0.000111\\
			&rand(0,1) 	&77.78	&0.000031	&100.00	&0.000101\\
			&rand(-1,1) 	&75.56	&0.000031	&100.00	&0.000104\\
			&xavier 	&88.89	&0.000033	&100.00	&0.000113\\
			&relu 	&100.00	&0.000042	&82.22	&0.000103\\\hline
			\multirow{5}{*}{BentIde}	&ortho 	&80.00	&0.000031	&95.56	&0.000101\\
			&rand(0,1) 	&86.67	&0.000032	&97.78	&0.000106\\
			&rand(-1,1) 	&97.78	&0.000033	&95.56	&0.000106\\
			&xavier 	&100.00	&0.000034	&100.00	&0.000117\\
			&relu 	&93.33	&0.000033	&97.78	&0.000132\\\hline
			\multirow{5}{*}{ArcTan}	&ortho 	&97.78	&0.000041	&97.78	&0.000104\\
			&rand(0,1) 	&80.00	&0.000036	&100.00	&0.000113\\
			&rand(-1,1) 	&100.00	&0.000038	&100.00	&0.000109\\
			&xavier 	&84.44	&0.000038	&100.00	&0.000121\\
			&relu 	&100.00	&0.000043	&97.78	&0.000126\\\hline
	\end{tabular}}
\end{table}
It is observed from TABLE \ref{iris_tab} that BF-ELM provides optimal accuracy for all combination of weight initialization scheme and activation function which shows the superior performance of BF-ELM to ELM. Further, studied are made with medium size and large complex applications.


\subsection{Benchmark with Satimage and Shuttle data-set}
\begin{figure}[H]
	\resizebox{0.49\textwidth}{!}{
		\begin{tikzpicture}
		\begin{axis}[
		grid=both,
		grid style={line width=.1pt, draw=gray!10},
		major grid style={line width=.2pt,draw=gray!50},
		minor tick num=5,
		scale only axis, 
		height=4cm,
		width=0.36\textwidth,
		xmax=101, xmin=0,
		ymax=101, ymin=0, 
		legend pos=north east,
		xtick={0,10,20,30,40,50,60,70,80,90,100}, 
		xlabel={\textbf{Number of nodes}},
		ylabel={\textbf{Accuracy}}, ylabel near ticks,
		legend style={
			at={(0.8,0.3)},
			anchor=north}
		]
		\addplot[blue, thick,mark=x] table [y=d,x=c,col sep=comma] {ResultsatImage.txt};
		\addplot[red, thick,mark=*] table [y=f,x=c,col sep=comma] {ResultsatImage.txt};
		\legend{ELM,BF-ELM}
		\end{axis}
		\end{tikzpicture}
	}
	\caption{Accuracy comparison of BF-ELM to ELM with respect to number of nodes on Sat-Image data-set }\label{satImage_node}
\end{figure}
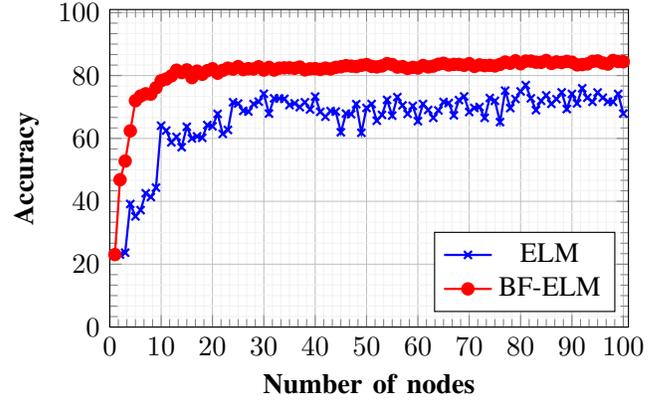

\begin{figure}[H]
	\resizebox{0.49\textwidth}{!}{
		\begin{tikzpicture}
		\begin{axis}[
		grid=both,
		grid style={line width=.1pt, draw=gray!10},
		major grid style={line width=.2pt,draw=gray!50},
		minor tick num=5,
		scale only axis, 
		height=4cm,
		width=0.36\textwidth,
		xmax=51, xmin=0,
		ymax=101, ymin=0, 
		legend pos=north east,
		xtick={0,10,20,30,40,50}, 
		xlabel={\textbf{Number of nodes}},
		ylabel={\textbf{Accuracy}}, ylabel near ticks,
		legend style={
			at={(0.8,0.3)},
			anchor=north}
		]
		\addplot[blue, thick,mark=x] table [y=d,x=c,col sep=comma] {ResultShuttel.txt};
		\addplot[red, thick,mark=*] table [y=f,x=c,col sep=comma] {ResultShuttel.txt};
		\legend{ELM,BF-ELM}
		\end{axis}
		\end{tikzpicture}
	}
	\caption{Accuracy comparison of BF-ELM to ELM w.r.t number of nodes on Shuttle data-set }\label{shuttle_node}
\end{figure}
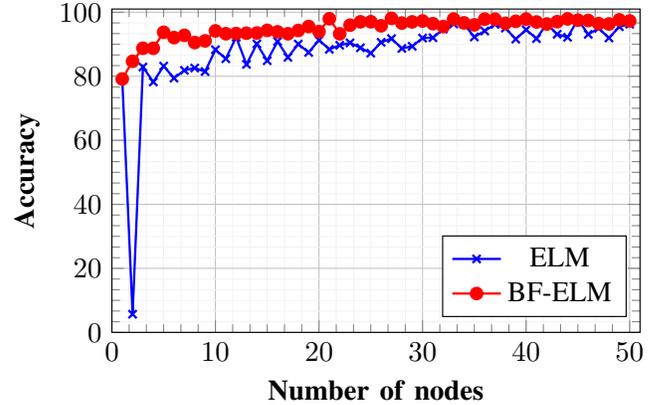

Satimage is one of medium size data-set having 4435 training and 2000 testing samples. The data-set contains 4 spectral band of $3\times 3$ neighborhood i.e. 36 predictive attributes for each sample and identified into seven classes namely red soil, cotton crop, gray soil, damp gray soil, soil with vegetation stubble, mixture class, and very damp gray soil. The testing accuracy with respect to number of hidden nodes is analyzed and summarized in Fig. \ref{satImage_node}. Similarly shuttle data-set consists of training sample count of 43,500 and testing size of 14,500 with nine attributes. The data-set have 7 classes namely Rad flow, Fpv close, Fpv open, High, Bypass, Bpv close, and Bpv open. The Fig. \ref{shuttle_node} depicts comparison of BF-ELM and ELM on testing set of shuttle data-set with respect to number of nodes.

The Fig. \ref{satImage_node} shows that BF-ELM converges to optimum accuracy with few nodes as compared to ELM. The testing accuracy obtained by BF-ELM is superior for every count of nodes up to 100 nodes. Further, testing is done with respect to variation of activation function and weight initialization scheme. The obtained results for 20 hidden nodes is given in TABLE  \ref{satimage_exp}.

The Fig. \ref{shuttle_node} shows superiority of BF-ELM over ELM in achieving testing accuracy. The summary of analysis over choice of weight initialization scheme and activation function is given in TABLE \ref{shuttle_exp}. 

\begin{table}[H]
	\renewcommand{\arraystretch}{1.3}
	\centering
	\caption{Accuracy comparison on Sat-Image test data-set with ELM for 20 hidden nodes}\label{satimage_exp}
	\resizebox{0.49\textwidth}{!}{\begin{tabular}{llllll}
			\hline
			\textbf{Activation} 	&	\textbf{Weight} 	&	\multicolumn{2}{c}{\textbf{ELM}} 	&	\multicolumn{2}{c}{\textbf{BF ELM}} 	\\\cline{3-6}
			\textbf{function}	&	\textbf{initialization}	&	\textbf{Acc}&\textbf{Test Time}	&	\textbf{Acc}&\textbf{Test Time}	\\\hline
			\multirow{5}{*}{None}	&ortho &63.95	&0.000407	&74.55	&0.000730\\
			&rand(0,1) &62.85	&0.000370	&74.45	&0.002063\\
			&rand(-1,1) &65.45	&0.000438	&74.60	&0.000654\\
			&xavier &62.60	&0.000372	&74.50	&0.001581\\
			&relu &63.45	&0.000391	&74.50	&0.000646\\\hline
			\multirow{5}{*}{Relu}	&ortho &66.95	&0.000572	&80.55	&0.002736\\
			&rand(0,1) &64.80	&0.000442	&80.60	&0.000827\\
			&rand(-1,1) &68.55	&0.000502	&75.35	&0.000970\\
			&xavier &68.25	&0.000500	&79.65	&0.000757\\
			&relu &63.60	&0.000513	&81.15	&0.002697\\\hline
			\multirow{5}{*}{Sigmoid}	&ortho &53.95	&0.000852	&83.05	&0.001171\\
			&rand(0,1) &23.05	&0.000541	&82.55	&0.001146\\
			&rand(-1,1) &47.95	&0.000844	&80.35	&0.000991\\
			&xavier &73.30	&0.000955	&82.20	&0.001273\\
			&relu &63.85	&0.002882	&82.00	&0.001010\\\hline
			\multirow{5}{*}{Tanh}	&ortho &65.25	&0.001031	&82.85	&0.001255\\
			&rand(0,1) &23.05	&0.000602	&81.10	&0.001398\\
			&rand(-1,1) &34.30	&0.000600	&80.15	&0.001204\\
			&xavier &62.35	&0.000898	&81.55	&0.001265\\
			&relu &39.00	&0.000629	&80.90	&0.001152\\\hline
			\multirow{5}{*}{Softsign}	&ortho &68.25	&0.001630	&81.25	&0.000740\\
			&rand(0,1) &75.20	&0.001567	&82.15	&0.000693\\
			&rand(-1,1) &75.35	&0.001555	&81.60	&0.000861\\
			&xavier &78.95	&0.000555	&81.20	&0.000992\\
			&relu &78.20	&0.000552	&81.30	&0.002090\\\hline
			\multirow{5}{*}{Sin}	&ortho &18.05	&0.000883	&81.10	&0.001281\\
			&rand(0,1) &18.30	&0.002236	&79.25	&0.000976\\
			&rand(-1,1) &19.35	&0.001087	&80.60	&0.001153\\
			&xavier &76.25	&0.002281	&82.35	&0.000971\\
			&relu &24.30	&0.001090	&79.35	&0.001147\\\hline
			\multirow{5}{*}{Cos}	&ortho &17.70	&0.001055	&80.65	&0.001890\\
			&rand(0,1) &17.10	&0.000940	&79.05	&0.002210\\
			&rand(-1,1) &17.40	&0.001169	&79.90	&0.001183\\
			&xavier &69.20	&0.000995	&80.10	&0.002716\\
			&relu &27.90	&0.000981	&81.60	&0.000988\\\hline
			\multirow{5}{*}{LeakyRelu}	&ortho &72.20	&0.000783	&82.05	&0.002142\\
			&rand(0,1) &63.10	&0.001910	&79.90	&0.000957\\
			&rand(-1,1) &67.75	&0.003062	&77.40	&0.001459\\
			&xavier &71.40	&0.002777	&80.40	&0.001300\\
			&relu &69.10	&0.002732	&80.35	&0.001703\\\hline
			\multirow{5}{*}{BentIde}	&ortho &69.70	&0.002916	&82.00	&0.000753\\
			&rand(0,1) &64.30	&0.001475	&81.85	&0.000944\\
			&rand(-1,1) &65.60	&0.000713	&82.25	&0.002872\\
			&xavier &71.80	&0.000621	&82.50	&0.000766\\
			&relu &62.85	&0.000561	&81.80	&0.000807\\\hline
			\multirow{5}{*}{ArcTan}	&ortho &70.65	&0.000779	&82.10	&0.000992\\
			&rand(0,1) &77.65	&0.000746	&83.35	&0.001195\\
			&rand(-1,1) &76.75	&0.000902	&81.20	&0.001165\\
			&xavier &78.60	&0.000882	&81.00	&0.000985\\
			&relu &76.55	&0.000851	&79.75	&0.001149\\\hline

	\end{tabular}}
\end{table}

\begin{table}[H]
	\renewcommand{\arraystretch}{1.3}
	\centering
	\caption{Accuracy comparison on Shuttle test data-set with ELM for 30 hidden nodes}\label{shuttle_exp}
	\resizebox{0.49\textwidth}{!}{\begin{tabular}{llllll}
			\hline
			\textbf{Activation} 	&	\textbf{Weight} 	&	\multicolumn{2}{c}{\textbf{ELM}} 	&	\multicolumn{2}{c}{\textbf{BF ELM}} 	\\\cline{3-6}
			\textbf{function}	&	\textbf{initialization}	&	\textbf{Acc}&\textbf{Test Time}	&	\textbf{Acc}&\textbf{Test Time}	\\\hline
			\multirow{5}{*}{None}	&ortho	&87.94	&0.001660	&87.94	&0.005870\\
			&rand(0,1)	&87.94	&0.001673	&87.94	&0.005970\\
			&rand(-1,1)	&87.94	&0.003528	&87.94	&0.002358\\
			&xavier	&87.94	&0.003402	&87.94	&0.006252\\
			&relu	&87.94	&0.003474	&87.94	&0.005913\\\hline
			\multirow{5}{*}{Relu}	&ortho	&92.88	&0.008845	&91.86	&0.005580\\
			&rand(0,1)	&88.56	&0.002351	&95.92	&0.005892\\
			&rand(-1,1)	&92.63	&0.007760	&94.24	&0.007834\\
			&xavier	&95.93	&0.003248	&96.61	&0.009495\\
			&relu	&92.71	&0.006785	&91.88	&0.003122\\\hline
			\multirow{5}{*}{Sigmoid}	&ortho	&95.83	&0.009473	&94.70	&0.005198\\
			&rand(0,1)	&79.15	&0.003522	&95.99	&0.006812\\
			&rand(-1,1)	&92.95	&0.004019	&98.66	&0.004478\\
			&xavier	&95.92	&0.004249	&96.54	&0.014445\\
			&relu	&93.32	&0.015859	&95.94	&0.014747\\\hline
			\multirow{5}{*}{Tanh}	&ortho	&92.02	&0.004684	&96.08	&0.004964\\
			&rand(0,1)	&79.16	&0.005102	&95.92	&0.004874\\
			&rand(-1,1)	&90.71	&0.009397	&96.11	&0.004620\\
			&xavier	&89.91	&0.005345	&97.00	&0.018385\\
			&relu	&92.28	&0.005576	&95.39	&0.017629\\\hline
			\multirow{5}{*}{Softsign}	&ortho	&91.70	&0.001886	&98.60	&0.006648\\
			&rand(0,1)	&79.11	&0.001870	&98.32	&0.007419\\
			&rand(-1,1)	&89.14	&0.001743	&95.81	&0.006926\\
			&xavier	&92.21	&0.004412	&94.92	&0.006698\\
			&relu	&91.47	&0.001895	&95.90	&0.002500\\\hline
			\multirow{5}{*}{Sin}	&ortho	&74.03	&0.004879	&94.22	&0.004050\\
			&rand(0,1)	&38.12	&0.004816	&95.30	&0.004491\\
			&rand(-1,1)	&40.59	&0.005819	&94.01	&0.004349\\
			&xavier	&76.89	&0.005369	&93.89	&0.004550\\
			&relu	&94.78	&0.005162	&94.88	&0.004606\\\hline
			\multirow{5}{*}{Cos}	&ortho	&67.63	&0.004991	&95.33	&0.004404\\
			&rand(0,1)	&33.88	&0.017163	&94.93	&0.004892\\
			&rand(-1,1)	&36.73	&0.004962	&94.49	&0.004495\\
			&xavier	&89.12	&0.004869	&94.78	&0.008545\\
			&relu	&96.68	&0.005868	&94.46	&0.005029\\\hline
			\multirow{5}{*}{LeakyRelu}	&ortho	&92.99	&0.014102	&92.19	&0.005792\\
			&rand(0,1)	&88.72	&0.008438	&92.94	&0.004327\\
			&rand(-1,1)	&93.44	&0.004335	&92.32	&0.013124\\
			&xavier	&93.92	&0.004438	&92.43	&0.012369\\
			&relu	&90.21	&0.012439	&93.19	&0.013407\\\hline
			\multirow{5}{*}{BentIde}	&ortho	&92.99	&0.006876	&95.72	&0.009425\\
			&rand(0,1)	&88.27	&0.007857	&96.29	&0.004027\\
			&rand(-1,1)	&92.90	&0.007002	&96.93	&0.003831\\
			&xavier	&94.16	&0.007719	&97.70	&0.008557\\
			&relu	&94.99	&0.007748	&97.35	&0.003863\\\hline
			\multirow{5}{*}{ArcTan}	&ortho	&91.48	&0.003868	&98.60	&0.007082\\
			&rand(0,1)	&79.13	&0.003311	&97.66	&0.012485\\
			&rand(-1,1)	&88.91	&0.003581	&97.97	&0.005168\\
			&xavier	&91.81	&0.006125	&96.91	&0.013778\\
			&relu	&92.80	&0.003609	&96.83	&0.005014\\\hline
	\end{tabular}}
\end{table}
It is observed from TABLE \ref{satimage_exp} that orthogonal initialization with sigmoid activation function and random initialization with arc-tan function gives best result for Satimage data-set. Similarly, TABLE \ref{shuttle_exp} shows that the best results are obtained with combination of random(-1,1) weight initialization with sigmoid activation function, orthogonal weights with soft-sign function, and orthogonal weights with arc-tan function achieves best performance score. The following sections depicts study that are carried out on large and complex data-sets.
\subsection{Benchmark with large forest cover data-set}

\begin{figure}[H]
	\resizebox{0.5\textwidth}{!}{\begin{tikzpicture}
		\begin{axis}[
		grid=both,
		grid style={line width=.1pt, draw=gray!10},
		major grid style={line width=.2pt,draw=gray!50},
		minor tick num=5,
		scale only axis, 
		height=4cm,
		width=0.36\textwidth,
		xmax=2001, xmin=0,
		ymax=70, ymin=50, 
		legend pos=north east,
		xtick={0,400,800,1200,1600,2000}, 
		xlabel={\textbf{Number of nodes}},
		ylabel={\textbf{Accuracy}}, ylabel near ticks,
		legend style={
			at={(0.8,0.3)},
			anchor=north}
		]
		\addplot[blue, thick,mark=x] table [y=d,x=c,col sep=comma] {ResultForest.txt};
		\addplot[red, thick,mark=*] table [y=f,x=c,col sep=comma] {ResultForest.txt};
		\legend{ELM,BF-ELM}
		\end{axis}
		\end{tikzpicture}}
	\caption{Accuracy comparison of BF-ELM to ELM w.r.t number of nodes on forest cover data-set}\label{Forest_node}
\end{figure}
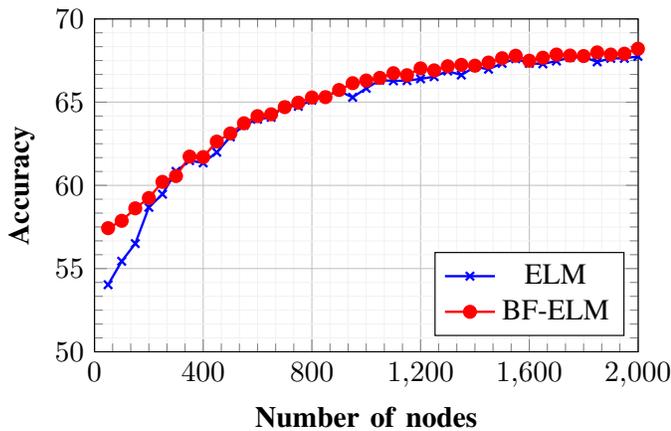

The proposed model is also tested for very large data-set of forest-cover type prediction application. The said data-set presents an extremely large prediction problem with seven classes. It contains 5,81,012 samples with 54 attributes randomly permuted over seven class namely spruce-fir, lodgepole pine, ponderosa pine, willow, aspen, doglas-fir and krummholz. The data-set is divided in to training and testing samples in accordance with the suggestion given in data-set description i.e. first 15,120 samples are used as training and rest 5,65,892 samples are used as testing. First experiment is conducted to study the effect of number of hidden nodes on both ELM and BF-ELM algorithms. The results obtained can be visualized in Fig. \ref{Forest_node}.
The Fig. \ref{Forest_node} shows that the accuracy on testing set increases for both ELM and BF-ELM algorithm. The figure depicts performance of both algorithms up to 2000 nodes and for each experiment conducted by increasing nodes, the accuracy obtained by BF-ELM is more than that of ELM. In second experiment the effect of weight initialization scheme and activation function is studied. For this experiment the number of nodes was set to 200 and orthogonal weight initialization scheme with sigmoid activation function is used for both algorithms. The obtained results are given in TABLE \ref{Forest_exp}. The table shows that for every combination of weight initialization scheme and activation function, the accuracy obtained by BF-ELM is superior to ELM. 

\begin{table}[H]
	\renewcommand{\arraystretch}{1.3}
	\centering
	\caption{Accuracy comparison on Forest cover data-set with ELM for 200 hidden nodes}\label{Forest_exp}
	\resizebox{0.49\textwidth}{!}{\begin{tabular}{llllll}
			\hline
			\textbf{Activation} 	&	\textbf{Weight} 	&	\multicolumn{2}{c}{\textbf{ELM}} 	&	\multicolumn{2}{c}{\textbf{BF ELM}} 	\\\cline{3-6}
			\textbf{function}	&	\textbf{initialization}	&	\textbf{Acc}&\textbf{Test Time}	&	\textbf{Acc}&\textbf{Test Time}	\\\hline
			\multirow{5}{*}{None}	&ortho	&54.28	&0.499435	&54.28	&0.622756\\
			&rand(0,1)	&54.27	&0.500526	&54.28	&0.511839\\
			&rand(-1,1)	&54.28	&0.471568	&54.28	&0.495267\\
			&xavier	&54.27	&0.470973	&54.28	&0.498806\\
			&relu	&54.27	&0.472855	&54.28	&0.502121\\\hline
			\multirow{5}{*}{Relu}	&ortho	&57.64	&0.990403	&57.16	&0.842121\\
			&rand(0,1)	&54.28	&0.598057	&56.99	&0.830652\\
			&rand(-1,1)	&57.52	&0.921646	&57.09	&0.868510\\
			&xavier	&57.26	&0.923671	&57.83	&0.811645\\
			&relu	&57.73	&0.921243	&56.81	&0.835724\\\hline
			\multirow{5}{*}{Sigmoid}	&ortho	&58.49	&0.754831	&59.77	&0.857107\\
			&rand(0,1)	&58.67	&0.987742	&58.97	&0.836054\\
			&rand(-1,1)	&58.88	&0.935462	&59.43	&0.840899\\
			&xavier	&58.03	&0.757011	&59.32	&0.834694\\
			&relu	&58.96	&0.742144	&59.58	&0.864178\\\hline
			\multirow{5}{*}{Tanh}	&ortho	&59.00	&1.021268	&59.47	&1.106261\\
			&rand(0,1)	&56.28	&1.303198	&59.45	&1.096017\\
			&rand(-1,1)	&57.97	&1.240969	&58.89	&1.126781\\
			&xavier	&57.88	&0.963619	&58.88	&1.113423\\
			&relu	&58.30	&0.957834	&59.77	&1.101439\\\hline
			\multirow{5}{*}{Softsign}	&ortho	&57.95	&0.579836	&60.13	&0.598289\\
			&rand(0,1)	&58.51	&0.561542	&59.19	&0.590835\\
			&rand(-1,1)	&58.43	&0.563180	&59.88	&0.588342\\
			&xavier	&58.38	&0.564881	&59.33	&0.587214\\
			&relu	&58.48	&0.562915	&59.09	&0.589696\\\hline
			\multirow{5}{*}{Sin}	&ortho	&58.42	&0.620325	&59.33	&0.645977\\
			&rand(0,1)	&57.94	&1.210180	&59.01	&0.646018\\
			&rand(-1,1)	&58.81	&0.798120	&59.01	&0.636448\\
			&xavier	&58.52	&0.613438	&59.26	&0.647153\\
			&relu	&58.14	&0.618676	&59.66	&0.652874\\\hline
			\multirow{5}{*}{Cos}	&ortho	&58.34	&0.648218	&59.14	&0.691882\\
			&rand(0,1)	&58.54	&1.240182	&59.82	&0.695169\\
			&rand(-1,1)	&58.82	&0.817066	&59.39	&0.686721\\
			&xavier	&59.15	&0.678123	&59.25	&0.701660\\
			&relu	&58.93	&0.645320	&59.46	&0.709892\\\hline
			\multirow{5}{*}{LeakyRelu}	&ortho	&58.80	&1.772426	&56.23	&1.278287\\
			&rand(0,1)	&54.27	&0.727510	&56.33	&1.349439\\
			&rand(-1,1)	&58.37	&1.620844	&57.27	&1.290102\\
			&xavier	&57.90	&1.603439	&56.90	&1.268446\\
			&relu	&58.07	&1.518761	&57.84	&1.304729\\\hline
			\multirow{5}{*}{BentIde}	&ortho	&58.56	&1.052638	&59.65	&1.071917\\
			&rand(0,1)	&59.08	&1.055782	&59.64	&1.076950\\
			&rand(-1,1)	&58.96	&1.049004	&59.77	&1.070770\\
			&xavier	&58.56	&1.049435	&58.81	&1.073052\\
			&relu	&59.00	&1.052567	&59.30	&1.079879\\\hline
			\multirow{5}{*}{ArcTan}	&ortho	&58.16	&0.711534	&60.49	&0.771294\\
			&rand(0,1)	&58.87	&0.912735	&58.98	&0.926173\\
			&rand(-1,1)	&58.19	&0.823203	&58.68	&0.786256\\
			&xavier	&58.22	&0.692458	&59.00	&0.845435\\
			&relu	&58.44	&0.673110	&59.68	&0.750680\\\hline
	\end{tabular}}
\end{table}
Further study are carried out on image data-sets, where the pixels are directly used as feature input to SLFN. This represents learning non-handcrafted feature directly from raw training images. The next two section presents performance study of MNIST and Brain-MRI data-set respectively.

\subsection{Benchmark with MNIST digit data-set}
\begin{figure}[H]
	\centering
	\includegraphics[width=0.5\textwidth]{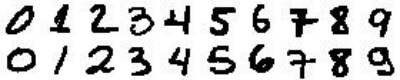}
	\caption{MNIST sample images}\label{mnist_img}
\end{figure}

Modified National Institute of Standards and Technology (MNIST) hand-written digit data-set is a standard for training and testing in the field of machine learning since 1999. The data-set consists of 60000 training and 10000 testing samples. The images have already been normalized to size $28 \times 28$ and presented in vector format. The Fig. \ref{mnist_img} shows some of the samples in MNIST data-set.\\

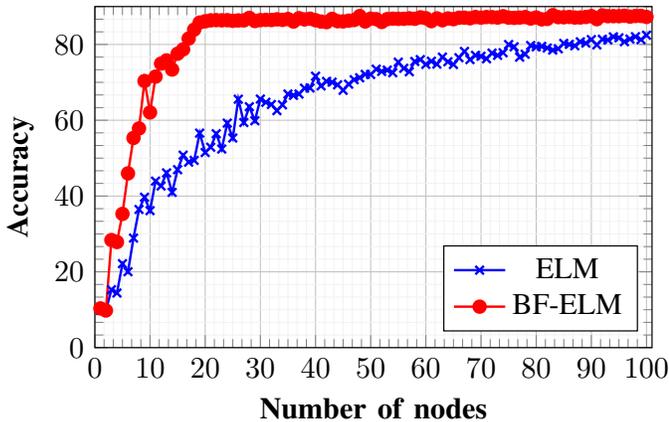
\begin{figure}[H]
	\resizebox{0.5\textwidth}{!}{\begin{tikzpicture}
		\begin{axis}[
		grid=both,
		grid style={line width=.1pt, draw=gray!10},
		major grid style={line width=.2pt,draw=gray!50},
		minor tick num=5,
		scale only axis, 
		height=4cm,
		width=0.36\textwidth,
		xmax=101, xmin=0,
		ymax=90, ymin=0, 
		legend pos=north east,
		xtick={0,10,20,30,40,50,60,70,80,90,100}, 
		xlabel={\textbf{Number of nodes}},
		ylabel={\textbf{Accuracy}}, ylabel near ticks,
		legend style={
			at={(0.8,0.3)},
			anchor=north}
		]
		\addplot[blue, thick,mark=x] table [y=d,x=c,col sep=comma] {Result1.txt};
		\addplot[red, thick,mark=*] table [y=e,x=c,col sep=comma] {Result1.txt};
		\legend{ELM,BF-ELM}
		\end{axis}
		\end{tikzpicture}}
	\caption{Accuracy comparison of BF-ELM to ELM w.r.t number of nodes on MNIST dataset}\label{MNIST_node}
\end{figure}

The first experiment is conducted to study the performance of ELM and BF-ELM with respect to number of nodes. The Fig. \ref{MNIST_node} represents the accuracy comparison of ELM and BF-ELM with orthogonal weight initialization and sigmoid activation function.

From Fig. \ref{MNIST_node} it is observed that BF-ELM achieves superior result with 20 hidden nodes and the testing accuracy keeps increasing with increase of hidden nodes. In second experiment the weight initialization and activation function are studied. The performance obtained for BF-ELM and ELM is represented in TABLE \ref{Mnist_exp1}. The experiment shows that BF-ELM achieves better accuracy in every combination. The best performance is achieved with xavier weight initialization and sigmoid activation function. The next experiment is carried out on pathological brain-MRI dataset which has gray level intensities in each image.

\begin{table}[H]
	\renewcommand{\arraystretch}{1.3}
	\centering
	\caption{Accuracy comparison on MNIST test data set with ELM for 20 hidden nodes}\label{Mnist_exp1}
	\resizebox{0.49\textwidth}{!}{\begin{tabular}{llllll}
		\hline
		\textbf{Activation} 	&	\textbf{Weight} 	&	\multicolumn{2}{c}{\textbf{ELM}} 	&	\multicolumn{2}{c}{\textbf{BF ELM}} 	\\\cline{3-6}
		\textbf{function}	&	\textbf{initialization}	&	\textbf{Acc}&\textbf{Test Time}	&	\textbf{Acc}&\textbf{Test Time}	\\\hline
		\multirow{5}{*}{None}	&ortho	&64.32	&0.068089	&86.02	&0.069997\\
		&rand(0,1)	&64.41	&0.069332	&85.97	&0.068331\\
		&rand(-1,1)	&61.37	&0.067602	&86.03	&0.068694\\
		&xavier	&60.07	&0.067313	&86.03	&0.068067\\
		&relu	&64.18	&0.068180	&85.99	&0.072153\\\hline
		\multirow{5}{*}{Relu}	&ortho	&54.71	&0.071155	&86.35	&0.070987\\
		&rand(0,1)	&64.90	&0.068160	&85.85	&0.068724\\
		&rand(-1,1)	&51.86	&0.069475	&85.73	&0.069322\\
		&xavier	&54.15	&0.068726	&85.52	&0.068729\\
		&relu	&55.00	&0.068493	&86.12	&0.070499\\\hline
		\multirow{5}{*}{Sigmoid}	&ortho	&64.35	&0.068715	&86.88	&0.069570\\
		&rand(0,1)	&11.33	&0.069876	&87.29	&0.069305\\
		&rand(-1,1)	&54.65	&0.068989	&86.65	&0.069786\\
		&xavier	&65.55	&0.068768	&87.56	&0.069762\\
		&relu	&63.36	&0.069838	&86.86	&0.069551\\\hline
		\multirow{5}{*}{Tanh}	&ortho	&64.07	&0.070599	&86.61	&0.070682\\
		&rand(0,1)	&11.35	&0.069093	&86.72	&0.070277\\
		&rand(-1,1)	&53.89	&0.070037	&86.94	&0.070810\\
		&xavier	&63.46	&0.070281	&86.95	&0.070624\\
		&relu	&58.14	&0.070568	&86.22	&0.072531\\\hline
		\multirow{5}{*}{Softsign}	&ortho	&62.51	&0.068898	&86.97	&0.069097\\
		&rand(0,1)	&62.58	&0.068350	&86.48	&0.069852\\
		&rand(-1,1)	&52.77	&0.069188	&87.10	&0.069221\\
		&xavier	&64.50	&0.067966	&87.14	&0.068566\\
		&relu	&63.53	&0.068463	&86.13	&0.069073\\\hline
		\multirow{5}{*}{Sin}	&ortho	&64.68	&0.068719	&86.98	&0.069508\\
		&rand(0,1)	&11.05	&0.071195	&86.60	&0.070107\\
		&rand(-1,1)	&13.67	&0.072125	&86.81	&0.069073\\
		&xavier	&61.21	&0.069393	&86.81	&0.069310\\
		&relu	&58.62	&0.069062	&86.87	&0.071229\\\hline
		\multirow{5}{*}{Cos}	&ortho	&45.94	&0.071003	&86.24	&0.068565\\
		&rand(0,1)	&11.10	&0.070971	&86.30	&0.069599\\
		&rand(-1,1)	&13.20& 0.070877	&86.46	&0.069696\\
		&xavier	&49.49	&0.068486	&86.72	&0.070056\\
		&relu	&50.85	&0.068768	&86.67	&0.069858\\\hline
		\multirow{5}{*}{LeakyRelu}	&ortho	&55.33	&0.070158	&86.29	&0.069612\\
		&rand(0,1)	&65.24	&0.068556	&86.58	&0.070003\\
		&rand(-1,1)	&50.45	&0.069813	&85.88	&0.071857\\
		&xavier	&50.64	&0.070322	&85.97	&0.070441\\
		&relu	&55.01	&0.070688	&86.94	&0.071169\\\hline
		\multirow{5}{*}{BentIde}	&ortho	&61.97	&0.068751	&86.42	&0.070512\\
		&rand(0,1)	&63.16	&0.068898	&86.23	&0.069749\\
		&rand(-1,1)	&60.36	&0.069847	&87.07	&0.071223\\
		&xavier	&62.28	&0.068689	&86.93	&0.070224\\
		&relu	&60.99	&0.069668	&86.44	&0.069452\\\hline
		\multirow{5}{*}{ArcTan}	&ortho	&61.83	&0.068867	&86.76	&0.070357\\
		&rand(0,1)	&60.23	&0.069516	&87.07	&0.070682\\
		&rand(-1,1)	&56.91	&0.070821	&86.91	&0.069662\\
		&xavier	&64.44	&0.069059	&86.92	&0.069933\\
		&relu	&63.51	&0.069914	&86.93	&0.071118\\\hline

	\end{tabular}}
\end{table}

\subsection{Benchmark with Multiclass brain MRI data-set:}

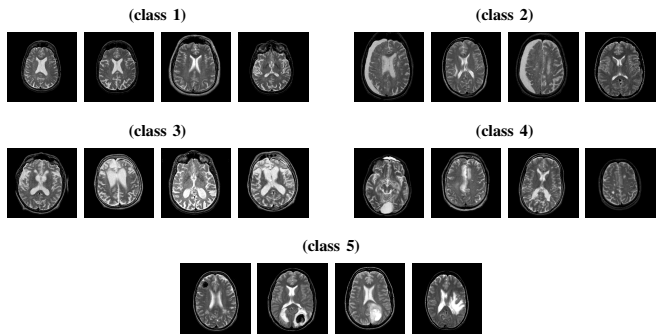
\begin{figure}[H]
	\centering
	\resizebox{0.49\textwidth}{!}{
		\begin{tikzpicture}
		\bf \large
		\node at (-2.5,1) {\includegraphics[height=1.8cm,width=1.8cm]{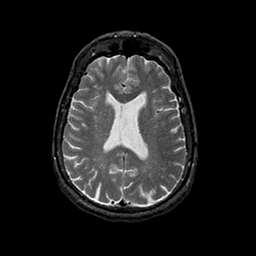}};
		\node at (-0.5,1) {\includegraphics[height=1.8cm,width=1.8cm]{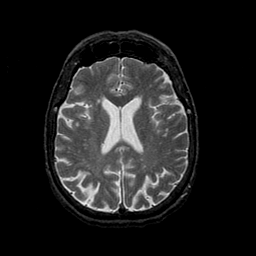}};
		\node at (1.5,1) {\includegraphics[height=1.8cm,width=1.8cm]{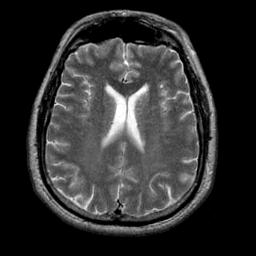}};
		\node at (3.5,1) {\includegraphics[height=1.8cm,width=1.8cm]{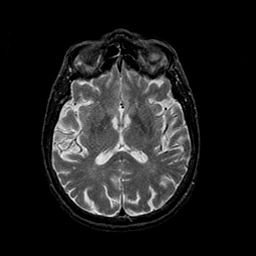}};
		
		\node at (6.5,1) {\includegraphics[height=1.8cm,width=1.8cm]{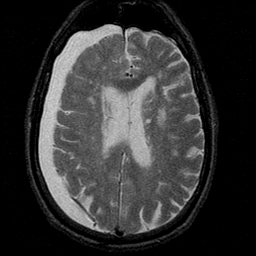}};
		\node at (8.5,1) {\includegraphics[height=1.8cm,width=1.8cm]{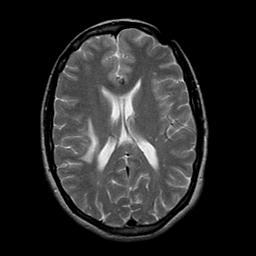}};
		\node at (10.5,1) {\includegraphics[height=1.8cm,width=1.8cm]{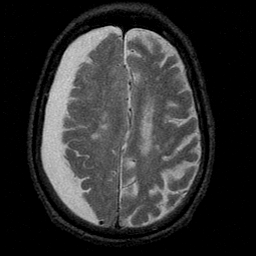}};
		\node at (12.5,1) {\includegraphics[height=1.8cm,width=1.8cm]{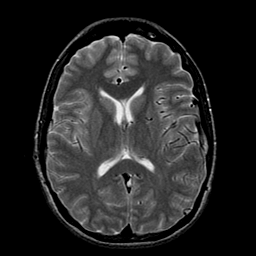}};
		
		\node at (-2.5,-2) {\includegraphics[height=1.8cm,width=1.8cm]{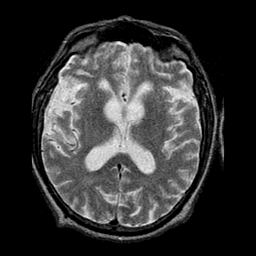}};
		\node at (-0.5,-2) {\includegraphics[height=1.8cm,width=1.8cm]{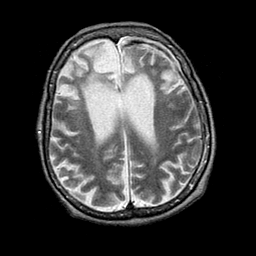}};
		\node at (1.5,-2) {\includegraphics[height=1.8cm,width=1.8cm]{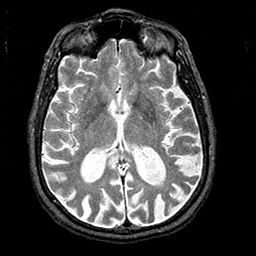}};
		\node at (3.5,-2) {\includegraphics[height=1.8cm,width=1.8cm]{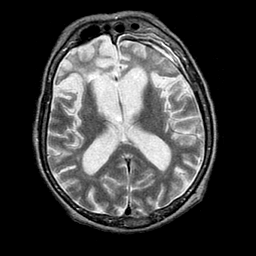}};
		
		\node at (6.5,-2) {\includegraphics[height=1.8cm,width=1.8cm]{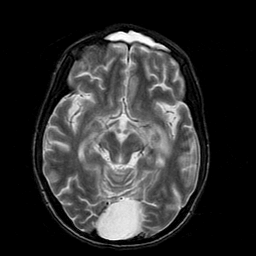}};
		\node at (8.5,-2) {\includegraphics[height=1.8cm,width=1.8cm]{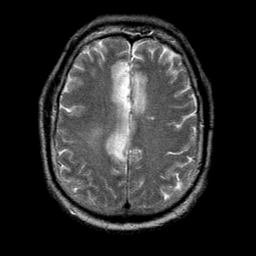}};
		\node at (10.5,-2) {\includegraphics[height=1.8cm,width=1.8cm]{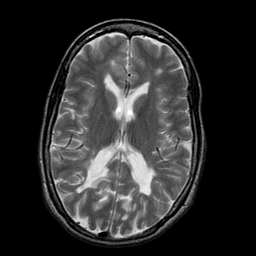}};
		\node at (12.5,-2) {\includegraphics[height=1.8cm,width=1.8cm]{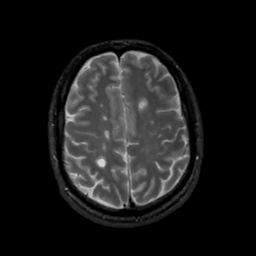}};
		
		\node at (2,-5) {\includegraphics[height=1.8cm,width=1.8cm]{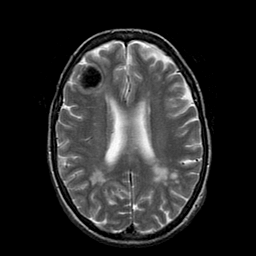}};
		\node at (4,-5) {\includegraphics[height=1.8cm,width=1.8cm]{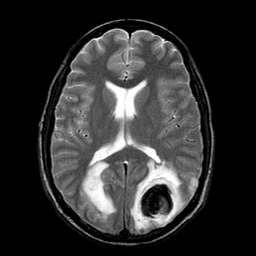}};
		\node at (6,-5) {\includegraphics[height=1.8cm,width=1.8cm]{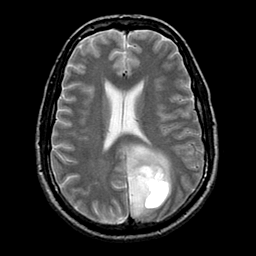}};
		\node at (8,-5) {\includegraphics[height=1.8cm,width=1.8cm]{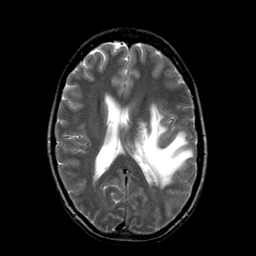}};
		
		\node[anchor=south] at (0.5,2) {(class 1)};
		\node[anchor=south] at (9.5,2) {(class 2)};
		\node[anchor=south] at (0.5,-1) {(class 3)};
		\node[anchor=south] at (9.5,-1) {(class 4)};
		\node[anchor=south] at (5,-4) {(class 5)};
		\end{tikzpicture}
	}
	\caption{Brain MRI samples}\label{brain-mri}
\end{figure}

The multiclass brain MR dataset comprises 200 images (40 normal and 160 pathological brain images) is used to evaluate the proposed model. The pathological brains contain diseases of four categories, namely brain stroke, degenerative, infectious and brain tumor; each category holds 40 images. The images are re-scaled to $80\times 80$ before applying to network directly. The Fig. \ref{brain-mri} shows some of the samples in brain-MRI dataset. The training and testing set is obtained by 80:20 stratified division.

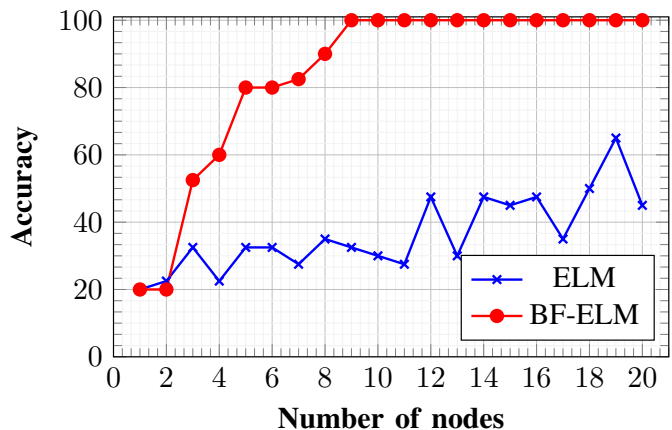
\begin{figure}[H]
	\resizebox{0.49\textwidth}{!}{\begin{tikzpicture}
		\begin{axis}[
		grid=both,
		grid style={line width=.1pt, draw=gray!10},
		major grid style={line width=.2pt,draw=gray!50},
		minor tick num=5,
		scale only axis, 
		height=4cm,
		width=0.36\textwidth,
		xmax=21, xmin=0,
		ymax=101, ymin=0, 
		legend pos=north east,
		xtick={0,2,4,6,8,10,12,14,16,18,20},
		xlabel={\textbf{Number of nodes}},
		ylabel={\textbf{Accuracy}}, ylabel near ticks,
		legend style={
			at={(0.8,0.3)},
			anchor=north}
		]
		\addplot[blue, thick,mark=x] table [y=d,x=c,col sep=comma] {Result2.txt};
		\addplot[red, thick,mark=*] table [y=e,x=c,col sep=comma] {Result2.txt};
		\legend{ELM,BF-ELM}
		\end{axis}
		\end{tikzpicture}}
	\caption{Accuracy comparison of BF-ELM to ELM w.r.t number of nodes on MRI (multi-class) dataset with (80:20) division}\label{MRI_node}
\end{figure}

The Fig. \ref{MRI_node} shows the results obtained during first experiment. In this the testing accuracy obtained by BF-ELM is compared to ELM with increasing number of hidden nodes up to 20. The experiment is carried out with orthogonal weight initialization scheme and sigmoid activation function. Here, it is observed that BF-ELM achieves best accuracy with 9 hidden nodes.

The results of second experiment is summarized in TABLE \ref{MRI_exp1} which depicts the effect of various weight initialization scheme and activation function for learning in SLFN using ELM and BF-ELM for 10 hidden nodes.

\begin{table}[H]
	\renewcommand{\arraystretch}{1.3}
	\centering
	\caption{Accuracy comparison on Brain MRI data set with ELM for 10 hidden nodes}\label{MRI_exp1} 
	\resizebox{0.49\textwidth}{!}{\begin{tabular}{llllll}
			\hline
			\textbf{Activation} 	&	\textbf{Weight} 	&	\multicolumn{2}{c}{\textbf{ELM}} 	&	\multicolumn{2}{c}{\textbf{BF ELM}} 	\\\cline{3-6}
			\textbf{function}	&	\textbf{initialization}	&	\textbf{Acc}&\textbf{Test Time}	&	\textbf{Acc}&\textbf{Test Time}	\\\hline
			\multirow{5}{*}{None}	&ortho	&57.50	&0.001152	&100.00	&0.001372\\
			&rand(0,1)	&52.50	&0.000811	&100.00	&0.000814\\
			&rand(-1,1)	&50.00	&0.000755	&100.00	&0.000893\\
			&xavier	&37.50	&0.000803	&100.00	&0.000791\\
			&relu	&47.50	&0.000705	&100.00	&0.000881\\\hline
			\multirow{5}{*}{Relu}	&ortho	&37.50	&0.000779	&100.00	&0.000885\\
			&rand(0,1)	&45.00	&0.000841	&95.00	&0.000909\\
			&rand(-1,1)	&57.50	&0.000775	&90.00	&0.000812\\
			&xavier	&32.50	&0.000770	&90.00	&0.000779\\
			&relu	&47.50	&0.000743	&87.50	&0.000767\\\hline
			\multirow{5}{*}{Sigmoid}	&ortho	&65.00	&0.000716	&100.00	&0.000926\\
			&rand(0,1)	&20.00	&0.000821	&100.00	&0.000862\\
			&rand(-1,1)	&42.50	&0.000775	&92.50	&0.000869\\
			&xavier	&35.00	&0.000722	&92.50	&0.000854\\
			&relu	&55.00	&0.000796	&95.00	&0.000892\\\hline
			\multirow{5}{*}{Tanh}	&ortho	&50.00	&0.000745	&97.50	&0.000939\\
			&rand(0,1)	&20.00	&0.000805	&77.50	&0.000856\\
			&rand(-1,1)	&27.50	&0.000724	&92.50	&0.000985\\
			&xavier	&50.00	&0.000832	&87.50	&0.000811\\
			&relu	&37.50	&0.000718	&57.50	&0.000854\\\hline
			\multirow{5}{*}{Softsign}	&ortho	&50.00	&0.000802	&95.00	&0.000929\\
			&rand(0,1)	&55.00	&0.001009	&92.50	&0.000770\\
			&rand(-1,1)	&37.50	&0.000752	&87.50	&0.000770\\
			&xavier	&47.50	&0.000732	&95.00	&0.000893\\
			&relu	&37.50	&0.000664	&95.00	&0.000770\\\hline
			\multirow{5}{*}{Sin}	&ortho	&17.50	&0.000701	&100.00	&0.000811\\
			&rand(0,1)	&20.00	&0.000872	&95.00	&0.000808\\
			&rand(-1,1)	&35.00	&0.000785	&100.00	&0.000811\\
			&xavier	&45.00	&0.000770	&95.00	&0.000863\\
			&relu	&32.50	&0.000948	&97.50	&0.000798\\\hline
			\multirow{5}{*}{Cos}	&ortho	&52.50	&0.000690	&100.00	&0.000817\\
			&rand(0,1)	&15.00	&0.000855	&97.50	&0.000836\\
			&rand(-1,1)	&27.50	&0.000719	&95.00	&0.000837\\
			&xavier	&62.50	&0.000805	&92.50	&0.000837\\
			&relu	&40.00	&0.000785	&95.00	&0.000817\\\hline
			\multirow{5}{*}{LeakyRelu}	&ortho	&57.50	&0.000691	&95.00	&0.000797\\
			&rand(0,1)	&42.50	&0.000697	&92.50	&0.000799\\
			&rand(-1,1)	&30.00	&0.000692	&100.00	&0.000799\\
			&xavier	&37.50	&0.000694	&97.50	&0.000831\\
			&relu	&47.50	&0.000901	&95.00	&0.000807\\\hline
			\multirow{5}{*}{BentIde}	&ortho	&37.50	&0.000722	&100.00	&0.000786\\
			&rand(0,1)	&52.50	&0.000836	&90.00	&0.000803\\
			&rand(-1,1)	&57.50	&0.000689	&100.00	&0.001013\\
			&xavier	&50.00	&0.000763	&100.00	&0.000786\\
			&relu	&52.50	&0.001027	&90.00	&0.000812\\\hline
			\multirow{5}{*}{ArcTan}	&ortho	&52.50	&0.000728	&100.00	&0.000816\\
			&rand(0,1)	&50.00	&0.000794	&100.00	&0.000821\\
			&rand(-1,1)	&50.00	&0.000706	&100.00	&0.000812\\
			&xavier	&55.00	&0.000723	&97.50	&0.000801\\
			&relu	&42.50	&0.000760	&97.50	&0.000797\\\hline

	\end{tabular}}
\end{table}

The above experiments highlight the performance improvement of SLFN learned by BF-ELM to SLFN learned by ELM. As there is only two pass in BF-ELM while ELM has one pass leaning, the proposed model takes twice the training time of ELM. However, the advantage BF-ELM is that the final network does not contain any random weights. Moreover, in many of the applications discussed above BF-ELM achieve better performance with less number of hidden nodes. 

\section{Conclusion}
This paper proposes a backward-forward algorithm for single hidden layer neural network which is a modified version of extreme learning machine. The proposed model performs better compared to ELM with fewer hidden nodes. Further, the evaluation of model with respect to weight various initialization scheme and activation functions proves the stability of the model as variance in the accuracy obtained for testing set is small compared to ELM. The proposed model can be directly used as classifier or can be used as a weight initialization model for fine tuning using gradient based model. In future, the model can be extended to multi layer neural network and convolutional neural network.


%

%
%
%
%
%
%
%



\bibliographystyle{IEEEtran}
\bibliography{Manuscript}

\begin{IEEEbiography}[{\includegraphics[width=1in,height=1.25in,clip,keepaspectratio]{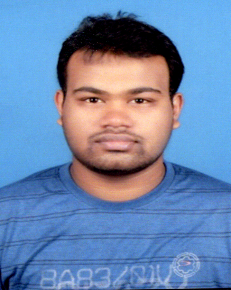}}]{Dibyasundar Das}
Dibyasundar Das is currently pursuing Ph. D in the Computer Science and Engineering at National Institute of Technology, Rourkela, India. He received his B.Tech degree in Information Technology from Biju Patnaik University of Technology, Rourkela, India, in 2011 and his M.Tech degree in Informatics from Siksha ‘O’ Anusandhan University, India in 2014. His current research interests include optical character recognition, pattern recognition and optimization. 
\end{IEEEbiography}
\begin{IEEEbiography}[{\includegraphics[width=1in,height=1.25in,clip,keepaspectratio]{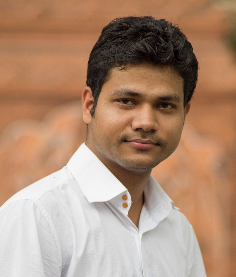}}]{Deepak Ranjan Nayak}
	Deepak Ranjan Nayak is currently with the Computer Science and Engineering at National Institute of Technology, Rourkela, India. His current research interests include medical image analysis, pattern recognition and cellular automata. He is currently serving as the reviewer of many reputed journals such as Multimedia Tools and Applications, IET Image Processing,  Computer Vision and Image Understanding, Computer and Electrical Engineering, Fractals, Journal of Medical Imaging and Health Informatics, IEEE Access, etc. He also serves as the reviewer of many conferences.
	
\end{IEEEbiography}
\begin{IEEEbiography}[{\includegraphics[width=1in,height=1.25in,clip,keepaspectratio]{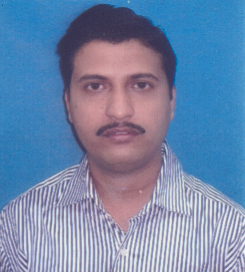}}]{Ratnakar Dash}
	Ratnakar Dash received his PhD degree from National Institute of Technology, Rourkela, India, in 2013. He is currently working as Assistant Professor in the Department of Computer Science and Engineering at National Institute of Technology, Rourkela, India. His field of interests include signal processing, image processing, intrusion detection system, steganography, etc. He is a professional member of IEEE, IE, and CSI. He has published forty research papers in journals and conferences of international repute. 
\end{IEEEbiography}
\begin{IEEEbiography}[{\includegraphics[width=1in,height=1.25in,clip,keepaspectratio]{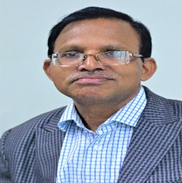}}]{Banshidhar Majhi}
	Banshidhar Majhi received his PhD degree from Sambalpur University, Odisha, India, in 2001. He is currently working as a Professor in the Department of Computer Science and Engineering at National Institute of Technology, Rourkela, India.  His field of interests include image processing, data compression, cryptography and security, parallel computing, soft computing, and biometrics. He is a professional member of MIEEE, FIETE, LMCSI, IUPRAI, and FIE. He serves as reviewer of many international journals and conferences. He is the author and co-author of over 80 journal papers of international repute. Besides, he has 100 conference papers and he holds 2 patents on his name. He has received “Samanta Chandra Sekhar Award” for the year 2016 by Odisha Bigyan Academy for his outstanding contributions to Engineering and Technology. 
\end{IEEEbiography}
%






\end{document}